\documentclass[conference]{IEEEtran}
\IEEEoverridecommandlockouts

\usepackage{cite}
\usepackage{amsmath,amssymb,amsfonts}
\usepackage{algorithmic}
\usepackage{graphicx}
\usepackage{textcomp}
\usepackage{xcolor}
\usepackage{booktabs,multirow}
\usepackage[linesnumbered,algoruled,boxed,lined]{algorithm2e}
\def\BibTeX{{\rm B\kern-.05em{\sc i\kern-.025em b}\kern-.08em
    T\kern-.1667em\lower.7ex\hbox{E}\kern-.125emX}}

\newcommand{\ie}{{\it i.e.}}
\newcommand{\eg}{{\it e.g.}}

\newcommand{\topN}{top-\emph{N}}

\begin{document}

\title{Collaborative Distillation for Top-$N$ Recommendation}

\author{\IEEEauthorblockN{Jae-woong Lee, Minjin Choi, Jongwuk Lee$^*$~\footnote{Corresponding author}}
\IEEEauthorblockA{
\textit{Sungkyunkwan University}\\
Republic of Korea \\
$\{$jwlee.icc, zxcvxd, jongwuklee$\}$@skku.edu}
\and
\IEEEauthorblockN{Hyunjung Shim}
\IEEEauthorblockA{
\textit{Yonsei University}\\
Republic of Korea\\
kateshim@yonsei.ac.kr}
}

\maketitle

\begin{abstract}
Knowledge distillation (KD) is a well-known method to reduce inference latency by compressing a cumbersome teacher model to a small student model. Despite the success of KD in the classification task, applying KD to recommender models is challenging due to the \emph{sparsity} of positive feedback, the \emph{ambiguity} of missing feedback, and the \emph{ranking} problem associated with the \topN\ recommendation. To address the issues, we propose a new KD model for the collaborative filtering approach, namely \emph{collaborative distillation} (\emph{CD}). Specifically, (1) we reformulate a loss function to deal with the ambiguity of missing feedback. (2) We exploit \emph{probabilistic rank-aware sampling} for the \topN\ recommendation. (3) To train the proposed model effectively, we develop two training strategies for the student model, called the \emph{teacher-} and the \emph{student-guided} training methods, selecting the most useful feedback from the teacher model. Via experimental results, we demonstrate that the proposed model outperforms the state-of-the-art method by 2.7-–33.2\% and 2.7-–29.1\% in hit rate (HR) and normalized discounted cumulative gain (NDCG), respectively. Moreover, the proposed model achieves the performance comparable to the teacher model.
\end{abstract}

\section{Introduction}

Neural recommender models~\cite{SedhainMSX15,WangWY15,KimPOLY16,HeLZNHC17,XueDZHC17,HeDWTTC18,TangW18a,WuDZE16,NiuCL18} have achieved better performance than conventional latent factor models either by capturing non-linear and complex correlation patterns among users/items, or by leveraging the hidden features extracted from auxiliary information such as texts and images. However, the number of model parameters of neural models is greater than that of conventional models by one or more orders of magnitude. This indicates a trade-off between accuracy and efficiency. As a result, neural recommender models usually suffer from higher latency during the inference phase.

Our primary goal is to develop a recommender model that achieves a balance between effectiveness and efficiency. In this paper, we employ \emph{knowledge distillation} (\emph{KD})~\cite{HintonVD15} which is a network compression technique by transferring the distilled knowledge of a large model (a.k.a., a \emph{teacher model}) to a small model (a.k.a., \emph{a student model}). As the student model can utilize the knowledge transferred from the teacher model, it naturally exhibits the properties of computational efficiency and low memory usage. Therefore, it is capable of achieving a balance between effectiveness and efficiency. 

Specifically, the training procedure for KD consists of two steps. In the offline training phase, the teacher model is supervised by a training dataset with labels. Then, the student model is learned to optimize two objectives: matching the label of a training sample (\ie, a \emph{hard target}) with that of model prediction and matching the label distribution (\ie, a \emph{soft target}) of the teacher model with that of the student model. In the inference phase, we utilize the student model. Because the teacher model possesses greater modeling power than the student model, the soft target serves as useful additional information for training the student model. The student model trained with KD can perform better than the student model only trained with the training set.

Despite the significant success of KD in the classification task, it is non-trivial to incorporate it into recommender models. More concretely, applying KD to recommender models involves several challenges: (1) Implicit user feedback is extremely sparse. (2) As users only provide positive feedback in implicit datasets, there is inherent ambiguity regarding unknown (or missing) feedback. That is, unknown feedback can be unlabeled positive or negative feedback. Such characteristics naturally require us to distinguish positive/negative feedback from unknown feedback. (3) Because a few top-ranked items are of interest to \topN\ recommendation, we should consider the degrees of importance of items based on their rankings.

Recently, Tang and Wang~\cite{TangW18b} proposed a KD model to address the ranking problem, called \emph{rank distillation} (\emph{RD}). RD uses only a few items with the highest rankings in the label distribution learned from the teacher model. Then, it manipulates them to positive feedback. In this sense, RD regards the knowledge transferred from the teacher model as augmented positive feedback, which helps alleviate the data sparsity problem associated with \topN\ recommendation.

Although RD improves the prediction accuracy of the student model, it is sub-optimal because some vital information in the soft target is ignored. First, the manipulation of the soft target in RD is only involved in generating additional positive feedback with the highest rankings. The key intuition of KD is that various correlations among items can provide additional information. In this regard, manipulating the soft target can distort the meaningful correlation patterns among items. Second, RD simply discards negative feedback with low rankings in the soft target. Removing low-ranked items from the soft target can make the process blind to negative user feedback. Therefore, both strategies in RD are counter-intuitive to the original idea of KD as they do not maintain the correlations among the items revealed in the soft target.

In this paper, we propose a new knowledge distillation model for collaborative filtering (CF), namely \emph{collaborative distillation} (\emph{CD}). Our model enjoys the advantages of both KD and RD. Specifically, the novelty of our model comes from the following aspects.

\vspace{1mm}
\noindent
\textbf{Reformulating a loss function for CF}. We design the CF model by revisiting the ambiguity of data representation. To resolve this issue, we propose a simple but improved CF loss function that only accounts for positive feedback. That is, unknown feedback is explicitly removed from the CF loss function. We claim that the presence of unknown feedback in the CF loss function can bias the prediction of ratings. As common implicit data representations treat unknown feedback as zero at all times, their predictions lean toward zero. By excluding unknown feedback from the CF loss function, we can prevent the prediction bias, thereby improving the overall performance of the student model.

\vspace{1mm}
\noindent
\textbf{Probabilistic rank-aware sampling}. Our model is influenced by the idea of RD, treating items differently based on their rankings. In the ranking problem, the higher-ranked items are more important because they can be potential inclusions in \topN\ recommendation. Therefore, we sample items in the soft target according to their rankings; the higher the ranking, the more the items are sampled. Because we sample both high- and low-ranked items in a probabilistic manner, our model can learn both positive/negative correlations among items. Therefore, we can take advantage of RD that considers the ranking order of items. Meanwhile, our method effectively overcomes the disadvantage of RD that ignores negative feedback among items. Besides, we rigorously preserve the idea of KD in the proposed model, because the probabilities in the soft target are used without manipulation. This enables us to fully exploit the correlations of items in the soft target. We believe that understanding the hidden correlations of items is crucial to overcome data sparsity and ambiguity problems.

\vspace{1mm}
\noindent
\textbf{Two training tactics in the student model}. Lastly, we develop two training tactics for the student model, called \emph{teacher-} and \emph{student-guided} methods. The teacher-guided method simply provides the soft target with the student model as in the conventional KD. In contrast, the student-guided method actively requests the useful items in the soft target to the teacher model by considering the training status of the student model. In other words, the student-guided method trains the student model by dynamically reflecting its status.

We conduct extensive experiments over the four benchmark datasets -- Amazon Music (AMusic), MovieLens 100k (ML100K), Yelp, and Gowalla. Through experimental results, we demonstrate that the proposed model significantly outperforms the state-of-the-art model (\ie, RD). Furthermore, the performance of the proposed model is comparable to that of the teacher model.
\section{Preliminaries}\label{sec:preliminaries}

In this section, we first introduce the basic notations and formulate the \topN\ recommendation problem. Then, we explain the concept of knowledge distillation (KD)~\cite{HintonVD15} and present rank distillation (RD)~\cite{TangW18b} that applies knowledge distillation to recommender models.

\vspace{1mm}
\noindent
\textbf{Problem statement}. For a set of users $\mathcal{U}=\{u_1, \dots, u_m\}$ and a set of items $\mathcal{I}=\{i_1, \dots, i_n\}$, we are given a user-item matrix $\mathbf{R} \in \{1,0\}^{m \times n}$, where $r_{ui} \in \mathbf{R}$ is the implicit user feedback represented by a binary (\ie, positive/negative) value assigned by user $u \in \mathcal{U}$ to item $i \in \mathcal{I}$. If $r_{ui} = 1$, it indicates \emph{known} (or \emph{observed}) feedback, implying positive user experience. Otherwise (\ie, $r_{ui} = 0$), it indicates \emph{missing} (or \emph{unobserved}) feedback, implying a mixture of unlabeled positive/negative preferences. Such ambiguity has been explicitly discussed in \emph{one-class collaborative filtering} (\emph{OCCF})~\cite{HuKV08,SindhwaniBHM10,PanZCLLSY08,PaquetK13,PanS09,LiHZC10,ZhengDMZ13,YaoTYXZSL14}. Given user $u$, $\mathcal{I}_u^+ = \{i \in \mathcal{I} | r_{ui} = 1\}$ is the set of items with known positive feedback, and $\mathcal{I}_u^- = \mathcal{I} \backslash \mathcal{I}_u^+$ is the set of items with missing feedback.

Our goal is to find a ranked list of the \topN\ items from implicit user feedback. Given user $u$, we need to rank the items (\ie, $i \in I_{u}^{-}$) according to their unknown preference scores. To achieve this goal, we define a ranking model $M(u, i; \theta)$ with a set of model parameters $\theta$ and compute a predicted preference score $\hat{r}_{ui}$ for each user $u$ and for each item $i$.

The ranking loss function can be categorized into three cases: \emph{point-wise}, \emph{pair-wise}, and \emph{list-wise}. In this paper, we focus on the point-wise loss, which is usually defined by the negative log likelihood of binary preference scores.
\begin{equation}
\begin{split}
\mathcal{L}(\theta) = - & \Big( \sum_{i \in \mathcal{I}_{u}^{+}}{\log{(P(r = 1 | u, i))}} \\
& + \sum_{i \in \mathcal{I}_{u}^{-}}{\log{(1 - P(r = 1 | u, i) \big)}} \Big),
\end{split}
\end{equation}

\noindent
where $\hat{r}_{ui} = P(r = 1 | u, i)$ represents the probability of item $i$ being preferred by user $u$.

\vspace{1mm}
\noindent
\textbf{Knowledge distillation (KD)}. This is a model-independent knowledge transfer framework designed to deliver the knowledge extracted from a complex teacher model to a simple student model. Many existing studies~\cite{ref03,ref04,ref09,ref10,ref11,ref12,ref13,ref23,ref26} have utilized KD to  compress deep neural networks as well as to achieve stable performances by distilling the knowledge from teacher model. In this process, KD makes use of a \emph{logit}, called a \emph{soft target}, which encodes the result values at the last layer before passing to the final activation layer. The success of KD can be attributed to the exploitation of hidden inter-class correlations in the soft target. Because the soft target reveals rich information, \ie, positive/negative correlations among items, the student model influenced by soft targets performs better than the same model trained only with ground-truth labels, called a \emph{hard target}.

The original KD and its variants are mainly developed in the context of the classification problem. They no longer remain valid in the \topN\ recommendation problem because of two reasons. First, many recommender models focus on solving the ranking problem; they aim to find the \topN\ items that the user most prefers. Although the label representation from both recommender and classification models is the same as the binary vector, the significance of each quantity is entirely different. In the classification model, both $0$ and $1$ are treated equally. Meanwhile, because the \topN\ recommendation model determines high-ranked items, it needs to place more weights on the responses to $1$.

Second, the recommender model needs to handle the ambiguity of missing feedback. Unlike the classificaton problem, missing feedback in the \topN\ recommendation problem can be either truly negative or hidden positive responses (\ie, preferred but unknown). When the teacher model regards all missing feedback as negative labels, the soft targets may be contaminated and be unable to bring for informative correlations between items. Consequently, the student model using the soft target may have worse performance than the original student model.

\begin{figure}
\includegraphics[width=\linewidth]{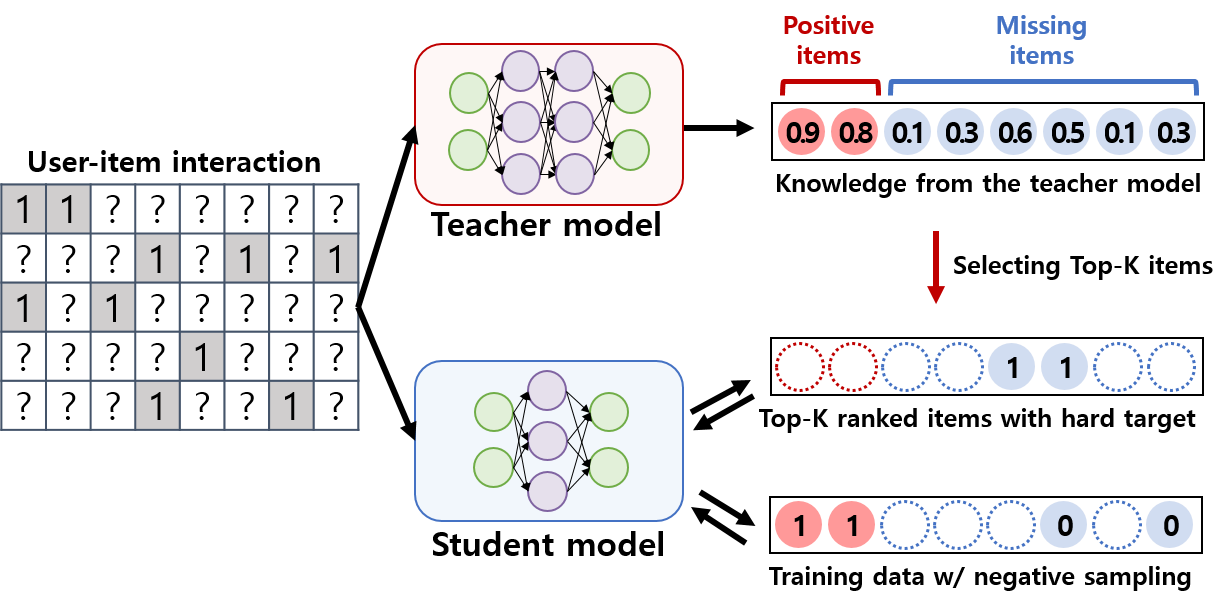}
\caption{Illustration of \emph{rank distillation} (RD)~\cite{TangW18b}. The teacher model transfers manipulated top-$k$ items as the distilled knowledge to the student model.}\label{fig:RD}
\vskip -0.2in
\end{figure}

\vspace{1mm}
\noindent
\textbf{Rank distillation (RD)}. Tang and Wang~\cite{TangW18b} proposed \emph{ranking distillation} (\emph{RD}) that applies KD for ranking models. As depicted in Fig.~\ref{fig:RD}, RD minimizes two losses: a \emph{ranking loss} with respect to the ground-truth ranking in the training dataset and a \emph{distillation loss} with respect to the top-$k$ ranking of unlabeled items.
\begin{equation}\label{eq:rd}
\mathcal{L}(\theta_S; \theta_T) = (1 - \rho)\mathcal{L}_{CF}(\theta_S) + \rho \mathcal{L}_{KD}(\theta_S; \theta_T),
\end{equation}

\noindent
where $\rho$ is the hyperparameter to balance two losses, and $\theta_S$ and $\theta_T$ are model parameters for the teacher and student models, respectively. $\mathcal{L}_{CF}(\cdot)$ is the ranking loss function for CF models and $\mathcal{L}_{KD}(\cdot)$ is the distillation loss function that guides the student model. For both ranking and distillation losses, RD employs the cross-entropy function using the negative log likelihood.
\begin{equation}\label{eq:cfloss1}
\begin{split}
\mathcal{L}_{CF}(\theta_S) = - & \Big( \sum_{i \in \mathcal{I}_{u}^{+}}{\log{(P(r = 1 | u, i))}} \\
& + \sum_{i \in \mathcal{I}_{u}^{-}}{\log{(1 - P(r = 1 | u, i) \big)}} \Big),
\end{split}
\end{equation}
\begin{equation}\label{eq:kdloss1}
\begin{split}
\mathcal{L}_{KD}(\theta_S; \theta_T) = -{ \sum_{i \in \{\pi_1, \dots, \pi_K\}}{w_i \log{\left(P(r = 1 | u, i; \theta_S, \theta_T)\right)}} },
\end{split}
\end{equation}
\noindent
where $P(r = 1 | u,i; \theta_S, \theta_T)$ is the preference probability of user $u$ for item $i$ in the soft target, $\{\pi_1, \dots, \pi_K\}$ indicates the top-$K$ items with missing feedback predicted by the teacher model, and $w_i$ is the importance weight to each item $i$.

Although RD addresses the ranking problem, it still suffers from several limitations. First, RD regards all missing feedback as negative feedback for the CF loss function in Equation~(\ref{eq:cfloss1}). Second, RD only makes use of top-$K$ ranked items in a deterministic manner, and merely ignores the other items in $\mathcal{I}_{u}^{-}$. As a result, the KD loss function in Equation~(\ref{eq:kdloss1}) includes no negative feedback. Lastly, RD modifies real-valued soft targets to positive feedback. The real-valued scores belong to the interval $[0, 1]$, and encode relative user preferences for items. Unfortunately, because RD quantizes all top-$K$ ranked items to positive feedback, it loses finer degrees of positive correlations among items learned from the teacher model.
\section{Proposed Model}\label{sec:model}

In this section, we propose a new knowledge distillation model for CF, namely \emph{collaborative distillation} (\emph{CD}) (Section~\ref{sec:overview}). Specifically, we first formulate a new CF loss function to address the ambiguity of missing feedback (Section~\ref{sec:CFloss}). Second, we design a probabilistic rank-aware sampling method to choose the items with unknown feedback (Section~\ref{sec:KDloss}). Lastly, we develop two training strategies for the student model, called teacher- and student-guided training strategies, to select the most beneficial label distributions from the teacher model (Section~\ref{sec:training}).

\subsection{Overview}\label{sec:overview}

\begin{figure}
\includegraphics[width=\linewidth]{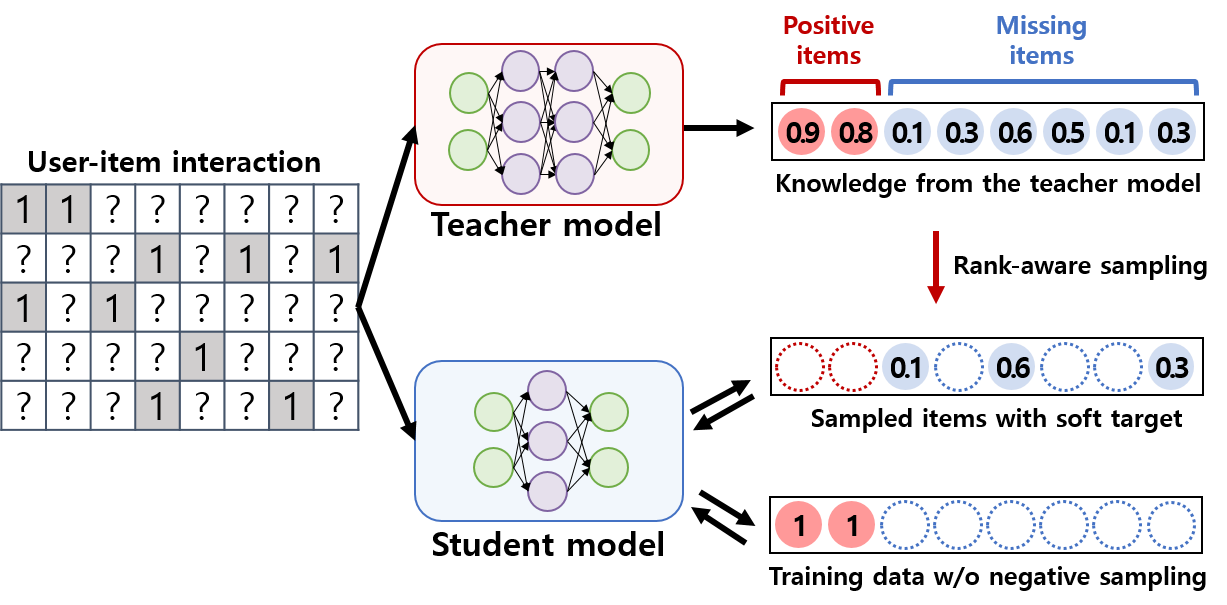}
\caption{Illustration of \emph{collaborative distillation} (CD).  The teacher model transfers the soft target to student model. We adopt probabilistic rank-aware sampling to place more weights for high-ranked items.}\label{fig:CD}
\vskip -0.2in
\end{figure}

Fig.~\ref{fig:CD} describes the overall procedure of our proposed model. Similar to the original KD, the training procedure of our model consists of two steps. First, we train a large teacher model $M(u, i; \theta_T)$ using a training dataset and its hard labels. The teacher model can be either a single CF model or an ensemble model combining multiple CF models. Once the training of the teacher model is completed, we can obtain the predicted preference scores for all unrated items in the teacher model. Then, we utilize both hard and soft targets for training a small student model $M(u, i; \theta_S, \theta_T)$.

In this process, we face the following challenges. In implicit datasets, user preferences are expressed in the form of positive/negative feedback. However, due to the nature of implicit data, we only observe sparse positive feedback. Besides, unlike the classification problem, the \topN\ recommendation is closely associated with the ranking problem. Therefore, the presence of a few \topN\ items among a block of unknown feedback should be further investigated.

To overcome these challenges, we first suggest an improved CF loss function, which considers the uncertainty in data representation of hard labels. Moreover, we utilize KD as a robust framework to address the ambiguity of unknown feedback. To this end, we extract two valuable pieces of information from the soft target of the teacher model: both positive/negative correlations among items and the candidates for high-ranked items. Our competitor, RD, manipulates the soft target to extract the rankings of items and uses it to define their KD loss function. As it is desirable that the ranking should reflect in the formulation of the CF, we place greater weight on higher-ranked items. We thus introduce a probabilistic rank-aware sampling method. It is interesting to note that RD has the drawback of sacrificing the precision of positive correlation and ignoring negative correlations between items. Unlike RD, we allow the ranking of items to reflect in KD and both positive and negative correlations to be utilized for the supervision of the student model.

Based on this idea, we reformulate the loss function for our model as a combination of the CF loss function and the KD loss function.
\begin{equation}\label{eq:kd}
\mathcal{L}(\theta_S; \theta_T) = \mathcal{L}_{CF}(\theta_S) + \lambda \mathcal{L}_{KD}(\theta_S; \theta_T).
\end{equation}

In the following sections, we explain the details of each loss function along with their design principle.

\subsection{Collaborative Filtering Loss}\label{sec:CFloss}

We design an improved collaborative filtering loss function to overcome the uncertainty of implicit data representation in CF. Both traditional KD and RD suggest computing the loss function from hard labels using the cross-entropy between the predicted label distribution and its hard label distribution. This method arguably treats all values of the label distribution equally importantly. However, positive feedback (\ie, $1$) indicates a preference, whereas negative feedback (\ie, $0$) can be either $1$ or $0$. We argue that, in terms of its confidence, the weight for $1$ should be much greater than the weight for $0$. Besides, treating all values equally in the cross-entropy loss function induces the model prediction to be biased toward $0$. To prevent this bias in prediction, we devise a selective cross-entropy loss function, which only matches items corresponding to $1$.
\begin{equation}\label{eq:cfloss2}
\mathcal{L}_{CF}(\theta_S) = - \sum_{i \in \mathcal{I}_{u}^{+}}{\log{\big(P(r = 1 | u, i)\big)}}.
\end{equation}

One might ask whether the method of predicting only $1$ can be applied to the training of the teacher model. Unfortunately, it always results in the prediction being $1$, which leads to a new bias. Thus, it is undesirable to apply the same strategy for the teacher model. Meanwhile, since the KD loss function provides positive/negative feedback for the training of the student model, the prediction is no longer biased in the proposed CF loss function. As a result, without the concern about training instability, the proposed CF loss function can provide more accurate feedback and eliminate the ambiguity of unknown feedback.

\subsection{Knowledge Distillation Loss}\label{sec:KDloss}

We devise a sampling-based KD loss function that not only distinguishes positive/negative feedback from missing feedback but also captures a user's relative preferences between items.
\begin{equation}\label{eq:kdloss2}
\begin{split}
\mathcal{L}_{KD}(\theta_S; \theta_T) = - & \Big( \sum_{i \in S(\mathcal{I}_{u}^{-})}{q_{ui}\log{\big(P(r = 1 | u, i)\big)}} \\
& + {(1 - q_{ui})\log{\big(1 - P(r = 1 | u, i)\big)}} \Big),
\end{split}
\end{equation}

\noindent
where $q_{ui}$ is a probability converted from the logit $z_{ui}$ and $S(\mathcal{I}_{u}^{-})$ is an item set sampled from $\mathcal{I}_{u}^{-}$. (We will explain how to compute $q_{ui}$ and how to identify $S(\mathcal{I}_{u}^{-})$ later.) The proposed KD loss function is different from that of RD in Equation~(\ref{eq:kdloss1}) in terms of two aspects. First, it utilizes the original soft target $q_{ui}$ just like the original KD. That is, it reflects both positive and negative correlations among items in the KD loss function. Second, the proposed loss function is computed by drawing the sampled items in a probabilistic manner. 

In this process, the sampling method is critical for the KD loss function. One baseline method can be a \emph{random sampling}, which learns the soft target regardless of the target values. Random sampling helps to reflect a user's relative preferences among different items. However, because it does not highlight the items with the highest rankings, it is inappropriate for \topN\ recommendation.

To explain our intuition, we first present several considerations: (1) Most of the items corresponding to unknown feedback represent negative preferences. Therefore, the randomly sampled items are likely to be biased toward negative preferences. (2) Although we can distinguish items with positive preferences from those with missing feedback, they should not be ranked higher than ones with known positive responses. Whereas the items with known positive feedback provide true positive experiences, the inferred positive feedback from the soft target might be incorrect or uncertain. (3) Items with positive scores in the soft target are likely to have positive correlations with the item for that soft target. Although the items with missing feedback might be uncertain, we believe that the soft target is still useful to capture relative preferences among items.

Based on these considerations, we develop a probabilistic rank-aware sampling method. The probability of sampling the item $i$ from $\mathcal{I}^-_{u}$ is determined by the ranking order among all unrated items, normalized by the total number of unrated items. We denote the importance of item $i$ by $\pi(i)$.
\begin{equation}
\pi(i) = \frac{rank(i)}{|\mathcal{I}^-_{u}|},
\end{equation}

\noindent
where $rank(i)$ is the relative ranking position of item $i$ in $\mathcal{I}^-_{u}$. That is, $rank(i) = 1$ denotes the highest ranking position and $rank(i) = |\mathcal{I}^-_{u}|$ denotes the lowest ranking position.

To draw a sample from unrated items, we investigate a rank-aware sampling function. First, we compute the sampling probability using a \emph{linear} function of the relative ranking position. To implement rank-aware linear sampling efficiently, we employ a rejection-based sampling method using rankings. 
\begin{equation}\label{eq:linear}
p_i \propto 1 - \pi(i).
\end{equation}
We can extend it to a non-linear sampling function using an \emph{exponential function}. With this adaptation, a few items with top-ranked positions have a much higher probability of being sampled than the others. The probabilities of the remaining items drop rapidly. This non-linear sampling function is formulated by:
\begin{equation}\label{eq:exponential}
p_i \propto \frac{1}{e^{\gamma \pi(i)}},
\end{equation}
\noindent
where $\gamma$ is the hyperparameter used to control the slope of the exponential function. The value is proportional to the gap between the top-ranked items and remaining ones.

Algorithm~\ref{alg:sampling} describes the pseudo-code for a rank-aware linear sampling method. In order to support exponential sampling, we can also modify uniform sampling (line 4). This sampling method is used in our proposed training strategies.

\SetKwInOut{Parameter}{Parameters}

\begin{algorithm}[t]
  \KwIn{Teacher model $M_T(u, i; \theta_T)$, unlabeled item set $\mathcal{I}_{u}^{-}$, sampling size $K$}
  
  \KwOut{Sampled item set $S(\mathcal{I}_{u}^{-})$ of size $K$}

  Compute all scores of unlabeled items in $\mathcal{I}_{u}^{-}$ using $M_T(u, i; \theta_T)$.

  Sort all unlabeled items by descending order of scores.
  
  \For{$i \in \mathcal{I}_{u}^{-}$}
  {
      Draw an item $j$ uniformly from $\mathcal{I}_{u}^{-}$ with replacement. \\
      
      \If{$rank(i) < rank(j)$}
      {
        $S(\mathcal{I}_{u}^{-}) \leftarrow \{i\} \cup S(\mathcal{I}_{u}^{-})$ \\
        
        \If{$|S(\mathcal{I}_{u}^{-})| = K$}
        {
            Break
        }
      }
  }
  
  \Return{$S(\mathcal{I}_{u}^{-})$}
  
\caption{Rank-aware linear sampling}\label{alg:sampling}
\end{algorithm}

\vspace{2mm}
\noindent
\textbf{Temperature in the KD loss}. One key factor of the original KD~\cite{HintonVD15} is to find a proper balance between the soft targets and hard labels. To tackle this issue,~\cite{HintonVD15} introduces the notion of a temperature $T$. Although the soft target is a useful resource for educating the student model, its distribution is often too sharp. In this case, because the relative correlation among items is not highlighted as much, the impact of KD is less significant.

When a softmax layer is the final output activation layer, it converts the logit $z_i$ to $q_i$ weighted by the temperature.
\begin{equation}
q_i = \frac{e^{z_i / T}}{\sum_{j}{e^{z_j / T}}},
\end{equation}
\noindent
where the temperature $T$ is directly proportional to the smoothness of the output probability distribution.

In this paper, we choose the point-wise approach for defining the user preference of items. Point-wise preferences are computed by a logistic function, which is a particular case of the softmax function. The logistic function is used to map a real-valued score to the probability of an item to be preferred ($r = 1$) as follows.
\begin{equation}
p_{ui} = P(r = 1 | u, i) = \frac{1}{1+e^{-z_{ui}}},
\end{equation}

\noindent
where $z_{ui}$ is the real-valued logit to user $u$ for item $i$.

Whereas the classification algorithm produces class probabilities using the softmax output layer, our problem can be regarded as a binary classification problem for each item. Then, the temperature for the logistic function is adopted using logits.
\begin{equation}
q_{ui} = P(r = 1 | u, i) = \frac{1}{1+e^{-\frac{z_{ui} + T_2}{T_1}}},
\end{equation}

\noindent
where $T_1$ and $T_2$ are the parameters for the temperature. $T_1$ controls the scale and $T_2$ controls the shift of $z_{ui}$. Although a more advanced function for temperature can be employed, we choose a relatively simple form of $T_1$ and $T_2$. We leave the formulation of various temperature functions as our future work.

\subsection{Interactive Training Tactics}\label{sec:training}

We investigate two training tactics for the student model: teacher-guided training and student-guided training. First, teacher-guided training is the process of learning from the teacher model, analogous to conventional KD training. The teacher model delivers the soft target to the student model without considering the training status of the student model. Then, the student model passively learns from those soft targets. 

In contrast, the student-guided training takes into account requests from the student model during training. That is, the student model examines its soft targets during training and draws several items according to the probabilistic rank-aware sampling method. Then, the student model asks the teacher model for the predictions (\ie, preference scores) of those items. In this way, the student model can instantly update the feedback of the high-ranked items from the teacher model. This idea is inspired by the idea of an interactive Q\&A section in the classroom, where students learn more efficiently by asking questions to their teachers. One may argue that the student model might not create meaningful questions during the early stage of its training. However, stupid questions are still better than random questions. Besides, student-guided training is advantageous because it helps to find an effective path for the training of the student model. We believe that this interactive training tactic helps om the fast convergence of model training and in approaching the improved solution via a better update path.

Note that the proposed training tactics are also used with our sampling method described in Section~\ref{sec:KDloss}. Even with the different sampling method, the adoption of the student-guided training is still suitable. That is, the student model can draw the items by any sampling method and then request for the feedback for those items from the teacher model. The process of each training tactic is as follows:

\vspace{1mm}
\noindent
\textbf{Teacher-guided training}. We choose the soft target from the ranking order in the teacher model. Therefore, the teacher model selects the sampled items without the intervention of the student model. Without any interaction between the two models, the student model learns the items selected by the teacher model.

\begin{figure}
\includegraphics[width=\linewidth]{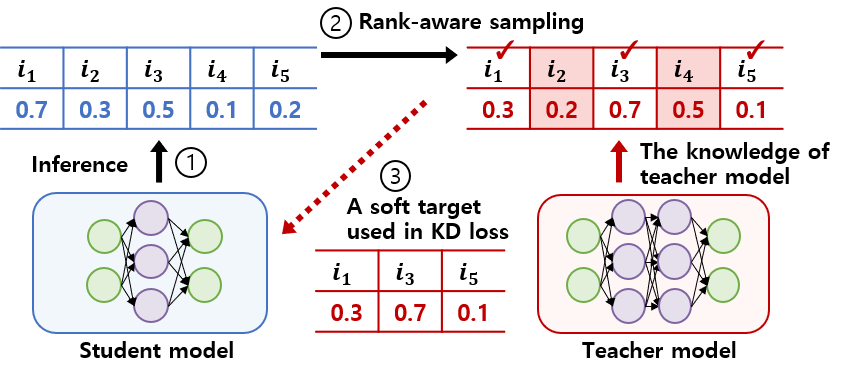}
\caption{Illustration of student-guided training. The student-guided training dynamically determines the teacher's soft target during student's inference.}\label{fig:SG}
\end{figure}

\vspace{1mm}
\noindent
\textbf{Student-guided training}. We choose the items from the soft target of the student model using the probabilistic rank-aware sampling method, as depicted in Fig.~\ref{fig:SG}. The student model dynamically selects the sampled items. Specifically, during the training of the student model, the soft target of the student model is analyzed. At each training step, some items are drawn to represent the soft target of the student model. Then, the student model asks the teacher model for predictions corresponding to those selected items. Based on this interaction, the KD loss function is updated. In Algorithm~\ref{alg:sampling}, we can replace the teacher model $M_T(u, i; \theta_T)$ with the student model $M_S(u, i; \theta_S)$ (line 1) to consider the interactive results between the teacher model and the student model at each training step.

\section{Experiments}\label{sec:experiments}

\subsection{Experimental Setup}\label{sec:setup}

\noindent
\textbf{Datasets}. We performed extensive experiments over the public benchmark datasets -- Amazon Music\footnote{http://jmcauley.ucsd.edu/data/amazon/}(AMusic), MovieLens 100K\footnote{https://grouplens.org/datasets/movielens/}(ML100K), Yelp\footnote{https://github.com/hexiangnan/sigir16-eals}, and Gowalla\footnote{http://dawenl.github.io/data/gowalla\_pro.zip}. We converted all ratings to a binary representation; either a user experiences an item positively or does not. These four datasets were selected to span over various degrees of data sparsity. Considering all the observed feedback as positive feedback, our goal was to identify the \topN\ recommendation for implicit datasets. As preprocessing, we filtered out the users who had less than 10 ratings and the items that were rated by less than 5 users. Table~\ref{tab:statistics} reports the detailed statistics of these datasets.

\begin{table*}[t!]
\begin{center}
\caption{Statistics of four benchmark datasets.}\label{tab:statistics}
\begin{tabular}{llllccc}
\toprule
Dataset & \# of users & \# of items & \# of interactions & Sparsity & min/max/avg. interactions per user & min/max/avg. interactions per item \\
\hline
AMusic & 2,831 & 13,410 & 63,054 & 99.83\% & 10/714/22.27 & 1/155/4.70 \\
ML100K & 943 & 1,682 & 100,000 & 93.70\% & 20/737/106.04 & 1/583/59.45 \\
Yelp & 9,788 & 25,373 & 489,820 & 99.80\% & 20/1024/50.04 & 1/674/19.30 \\
Gowalla & 13,149 & 14,009 & 535,650 & 99.71\% & 15/764/40.73 & 15/1743/38.24 \\
\bottomrule
\end{tabular}
\end{center}
\vskip -0.2in
\end{table*}

\begin{table*}[t]
\caption{Performance comparison of each model in four benchmark datasets.}\label{tab:com1}
\begin{center}
\begin{tabular}{p{1cm}p{1.1cm}p{1.1cm}p{1.5cm}p{1.1cm}p{1.5cm}p{1.1cm}p{1.5cm}p{1.1cm}p{1.5cm}}
\toprule
\multicolumn{2}{c}{Models} & \multicolumn{2}{c}{AMusic} & \multicolumn{2}{c}{ML100K} & \multicolumn{2}{c}{Yelp} & \multicolumn{2}{c}{Gowalla}\\
 &  & HR@50 & NDCG@50 & HR@50 & NDCG@50 & HR@50 & NDCG@50 & HR@50 & NDCG@50  \\
\hline
\multirow{7}{*}{CDAE}
& Teacher & 0.1727 & 0.0547 & 0.3917 & 0.1288 &0.1150 & 0.0340 & 0.3057 & 0.1269 \\
& Student & 0.1217 & 0.0370 & 0.3565 & 0.1107 &0.0956 & 0.0278 & 0.2632 & 0.1088 \\
& RD & 0.1275 & 0.0402 & 0.3578 & 0.1112 & 0.0949 & 0.0272 & 0.2638 & 0.1092 \\
& RD-Rank & 0.1238 & 0.0366 & 0.3580 & 0.1110 & 0.0915 & 0.0258 & 0.2602 & 0.1034 \\
& CD-Base & 0.1613 & 0.0498 & 0.3707 & 0.1124 & 0.1042 & 0.0304 & 0.2613 & 0.1093 \\
& CD-TG & \text{0.1653} & \text{0.0513} & \textbf{0.3805} & \text{0.1175} &\text{0.1060} & \text{0.0309} & \text{0.2682} & \text{0.1113} \\
& CD-SG & \textbf{0.1681} & \textbf{0.0519} & \text{0.3741} & \textbf{0.1182} & \textbf{0.1067} & \textbf{0.0313} & \textbf{0.2710} & \textbf{0.1122}\\
\cline{2-10}
& Gain (\%) & 31.8 & 29.1 & 6.3 & 6.3 & 12.4 & 15.1 & 2.7 & 2.7\\
\hline
\multirow{7}{*}{Caser}
& Teacher & 0.1366 & 0.0392 & 0.3145 & 0.0868 & 0.0947 & 0.0266 & 0.3005 & 0.1109 \\
& Student & 0.0919 & 0.0276 & 0.2717 & 0.0730 & 0.0789 & 0.0220 & 0.2033 & 0.0768 \\
& RD & 0.0936 & 0.0276 & 0.2732 & 0.0758 & 0.0814 & 0.0230 & 0.2358 & 0.0877 \\
& RD-Rank & 0.0909 & 0.0271 & 0.2787 & 0.0774 & 0.0813 & 0.0232 & 0.2362 & 0.0880\\
& CD-Base & 0.1211 & \textbf{0.0355} & 0.3147 & 0.0872 & 0.0874 & 0.0244 & 0.2557 & \textbf{0.0943} \\
& CD-TG & 0.1135 & 0.0336 & \textbf{0.3203} & \text{0.0879} & 0.0899 & 0.0249 & \text{0.2525} & \text{0.0904} \\
& CD-SG & \textbf{0.1247} & \text{0.0351} & 0.3196 & \textbf{0.0891} & \textbf{0.0965} & \textbf{0.0269} & \textbf{0.2570} & 0.0925 \\
\cline{2-10}
& Gain (\%) & 33.2 & 28.6 & 17.2 & 17.5 & 18.6 & 17.0 & 9.0 & 7.5\\
\hline
\bottomrule
\end{tabular}
\end{center}
\vskip -0.2in
\end{table*}


\vspace{2mm}
\noindent
\textbf{Competitive models}. Since RD~\cite{TangW18b} is the state-of-the-art KD model for \topN\ recommendation, we compare the proposed model with the original RD. Besides, we evaluated two baseline models, RD-Rank and CD-Base, modifying the sampling method for RD and CD, respectively. We also validate the effect of our rank-aware sampling in RD and CD.

\begin{itemize}
\item \textbf{RD}~\cite{TangW18b}: To define the KD loss in equation~(\ref{eq:kdloss1}), this utilizes only the top-$K$ items of the soft target by quantizing their values to $1$.

\item \textbf{RD-Rank}: This employs the same loss function for RD as in Equation~(\ref{eq:rd}), but selects the items for the KD loss function in Equation~(\ref{eq:kdloss1}) using rank-aware sampling.

\item \textbf{CD-Base}: This is our proposed model using CD as in Equation~(\ref{eq:kd}) but selects the items for the KD loss in Equation~(\ref{eq:kdloss2}) using random sampling.

\item \textbf{CD-TG}: This is our proposed model using teacher-guided model training.

\item \textbf{CD-SG}: This is our proposed model using student-guided model training.
\end{itemize}

To validate the proposed model, we chose two state-of-the-art recommender models-- CDAE~\cite{WuDZE16} and Caser~\cite{TangW18a}. (This paper focuses on \topN\ recommender models with point-wise preferences. We leave the evaluation for other models with pair-wise preferences, \eg, NPR~\cite{NiuCL18}, to future work.) Although the teacher model can be an ensemble model combining multiple models, we focused on verifying our model for the simple case. Finally, the same recommender models having high complexity and low complexity were chosen for the teacher and the student model, respectively. Note that this setting is consistent with existing KD studies.


\vspace{2mm}
\noindent
\textbf{Evaluation protocol}. We adopted the \emph{leave-one-out} evaluation~\cite{HeLZNHC17,XueDZHC17,HeDWTTC18}. Specifically, we held-out the last timestamp user-item interaction as the test data for each user, and the rest of user-item interactions are used for training data. Unlike sampling-based evaluation~\cite{HeLZNHC17,XueDZHC17,HeDWTTC18} that randomly chose 100 items from the set of unrated items, we chose all unrated items as the candidate items. We believe that this evaluation protocol is time-consuming but more thorough.

\vspace{2mm}
\noindent
\textbf{Evaluation metrics}. We measured the accuracy of \topN\ recommendation for two metrics, hit rate (HR) and normalized discounted cumulative gain (NDCG), as done in existing studies~\cite{HeLZNHC17,XueDZHC17,HeDWTTC18}. The size $N$ of the ranked list was chosen to be 50 for HR@N and NDCG@N, respectively. HR@N examines whether or not the test item is present in the \topN\ list, and NDCG@N places more weights on higher-ranked items than others in the \topN\ list. In both metrics, the value is proportional to the accuracy of the result. Both metrics are averaged across all users.

\vspace{2mm}
\noindent
\textbf{Reproducibility}. To ensure a fair evaluation, each hyperparameter and regularization term was fine-tuned and shared among all KD models. We randomly initialized model parameters using Gaussian distribution $\mathcal{N}(0, 1)$. Specifically, each baseline CF model had the following hyperparameters.

\begin{itemize}
    \item CDAE~\cite{WuDZE16}: The latent dimensions for the teacher and the student model were 100 and 10, respectively. We set the number of negative sampling items to be 0.5*$\mathcal|{I}_{u}|$ and the denoising ratio as 0.1. We used the Adagrad optimizer with learning rate = 0.2, l2-regularizer = 0.001, and batch size = 256.

    \item Caser~\cite{TangW18a}: The latent dimensions for the teacher and the student model were 50 and 5, respectively. We set sequence length $L$ to be 5, target length $T$ to be 1, and the number of negative sampling items to be $3$. We used the Adam optimizer with learning rate = 0.001, l2-regularizer = 0.000001, dropout ratio = 0.5, and batch size = 512.
\end{itemize}

\begin{figure*}[t]
\centering
\begin{tabular}{cccc}
\includegraphics[width=0.23\textwidth]{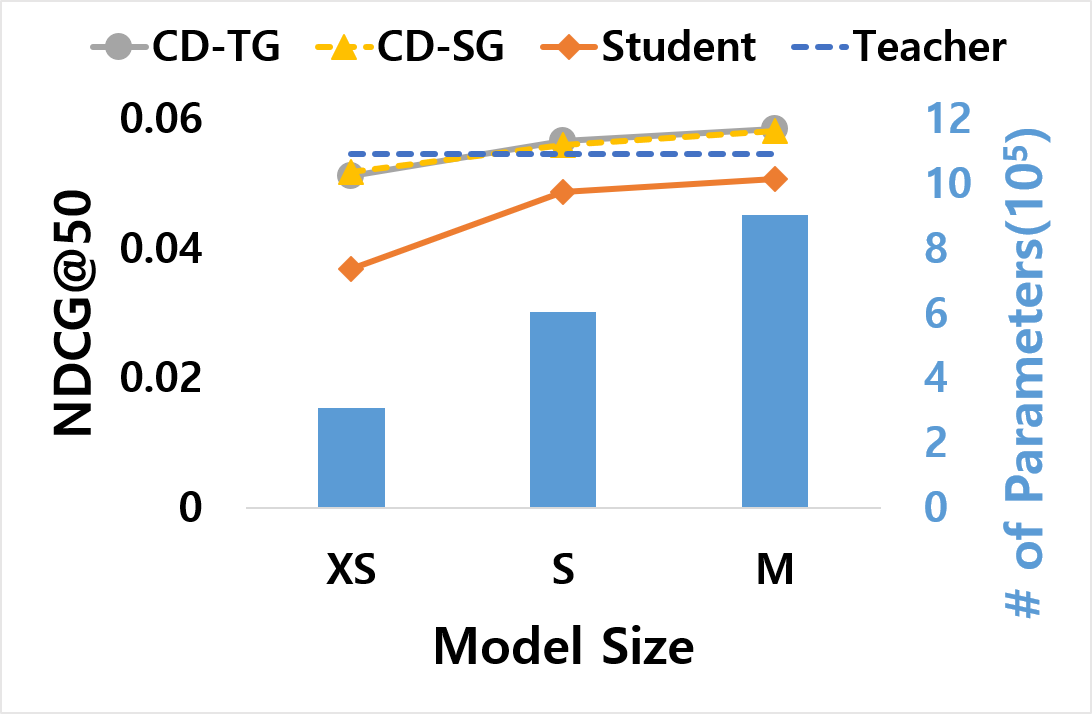} &
\includegraphics[width=0.23\textwidth]{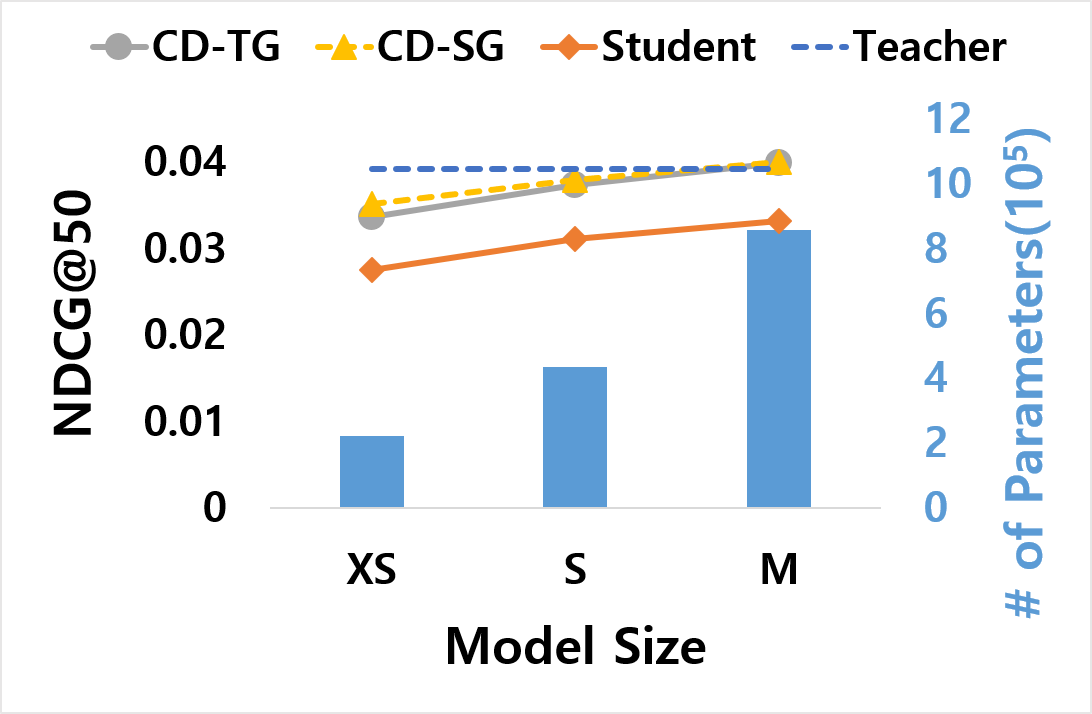} &
\includegraphics[width=0.23\textwidth]{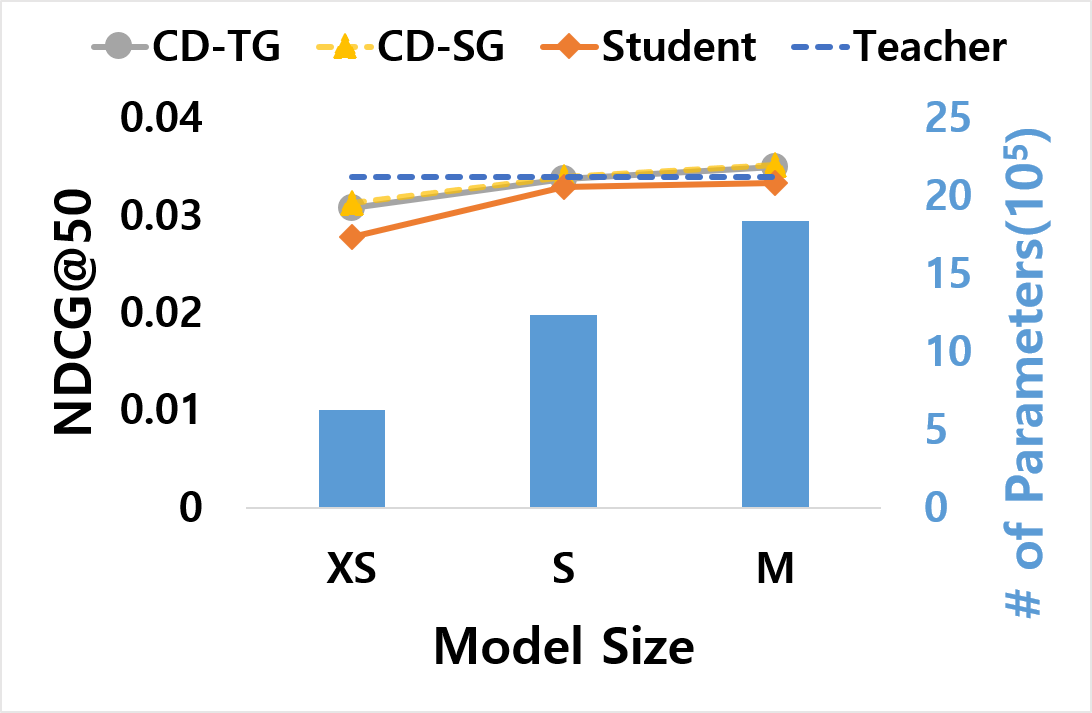} &
\includegraphics[width=0.23\textwidth]{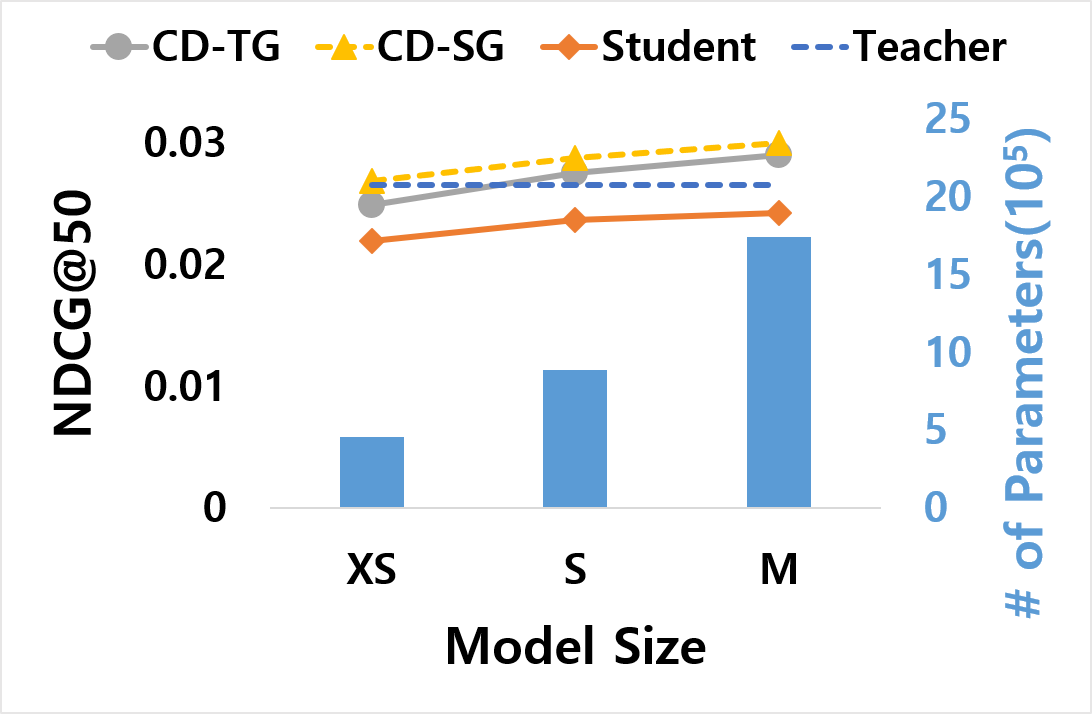} \\
(a) AMusic (CDAE) & (b) AMusic (Caser) & (c) Yelp (CDAE) & (d) Yelp (Caser) \\
\end{tabular}
\caption{NDCG@50 across three different model sizes. XS, S, and M indicate 10\%, 20\%, 30\% of the size of teacher model, respectively. }\label{fig:modelvseffectiveness}
\end{figure*}

\begin{figure*}[t]
\centering
\begin{tabular}{cccc}
\includegraphics[width=0.23\textwidth]{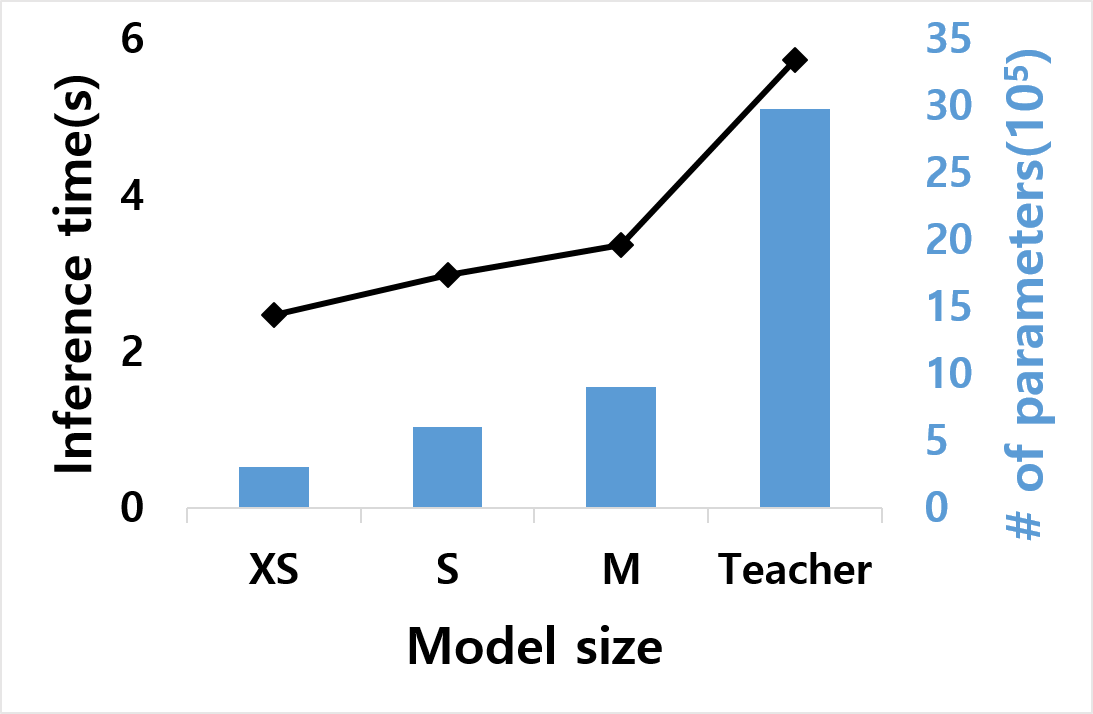} &
\includegraphics[width=0.23\textwidth]{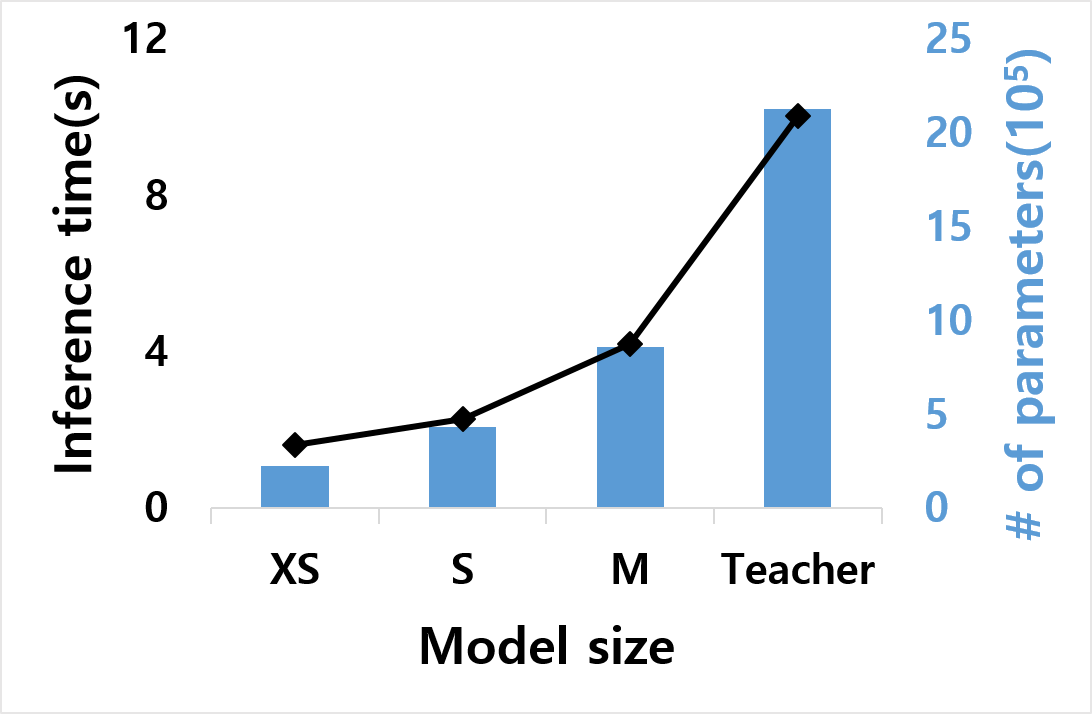} &
\includegraphics[width=0.23\textwidth]{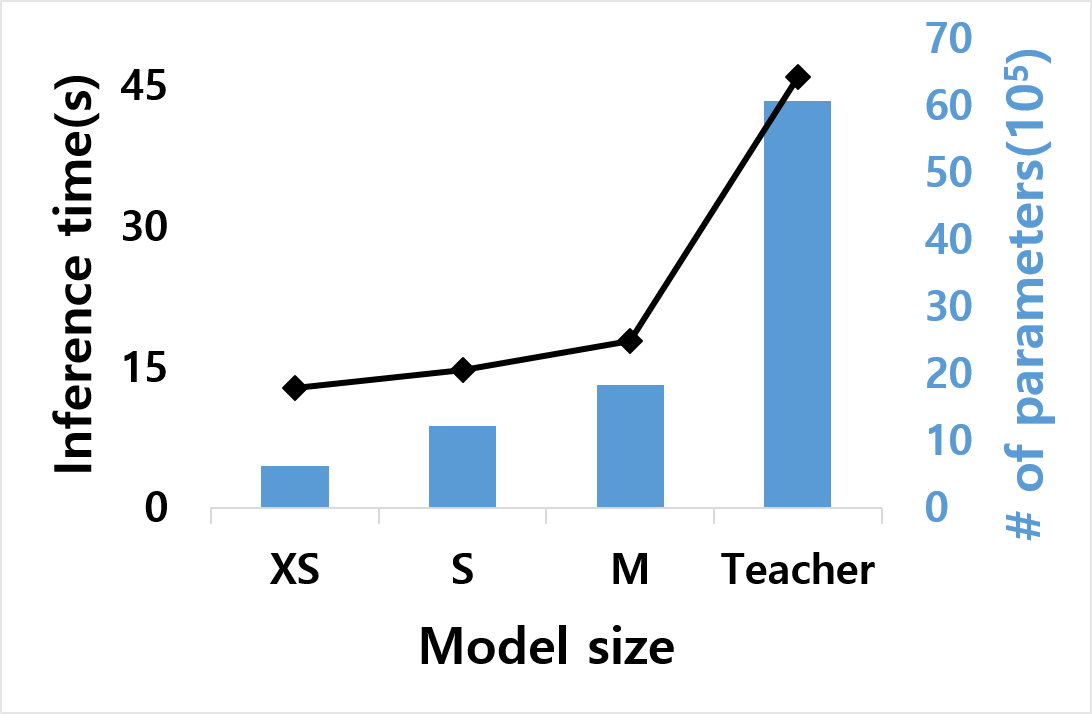} &
\includegraphics[width=0.23\textwidth]{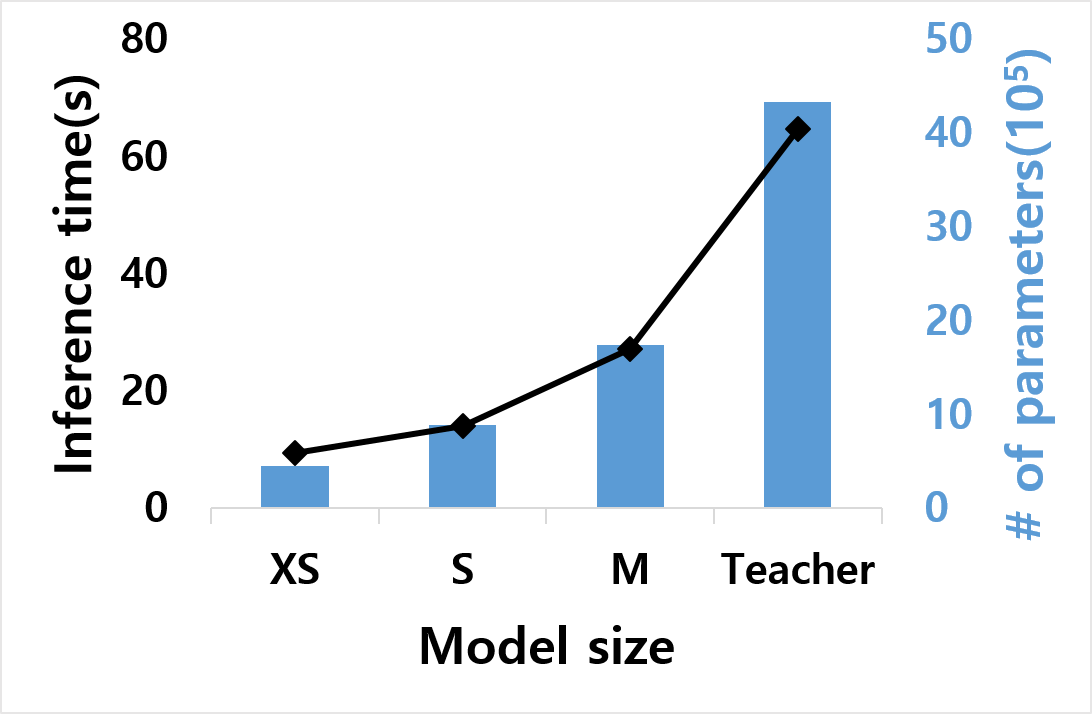} \\
(a) AMusic (CDAE) & (b) AMusic (Caser) & (c) Yelp (CDAE) & (d) Yelp (Caser) \\
\end{tabular}
\caption{Online inference latency vs. model size. In both datasets, the inference latency is greatly reduced by decreasing the model parameters.}\label{fig:modelvsinference}
\end{figure*}

\begin{figure*}[t]
\centering
\begin{tabular}{cccc}
\includegraphics[width=0.23\textwidth]{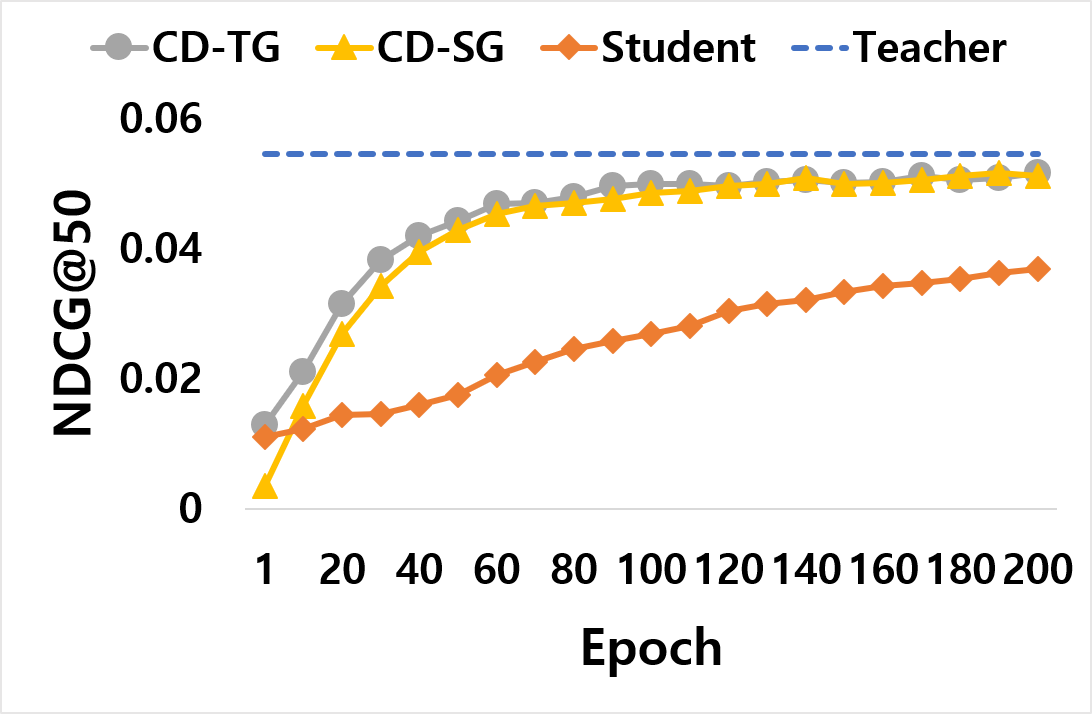} &
\includegraphics[width=0.23\textwidth]{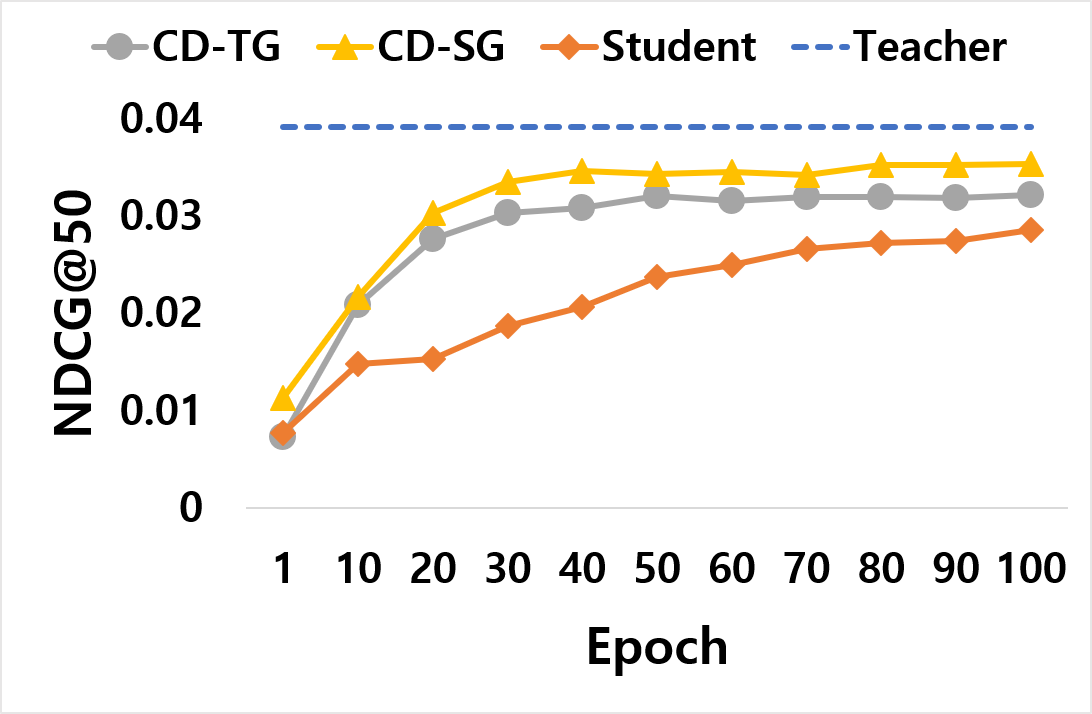} &
\includegraphics[width=0.23\textwidth]{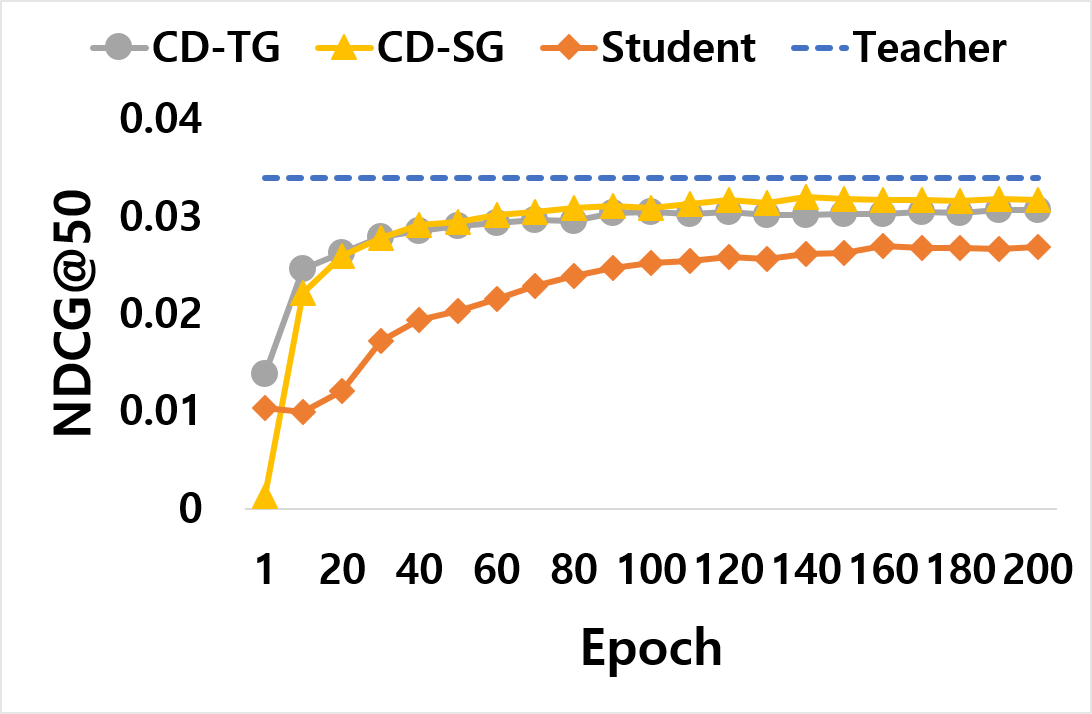} &
\includegraphics[width=0.23\textwidth]{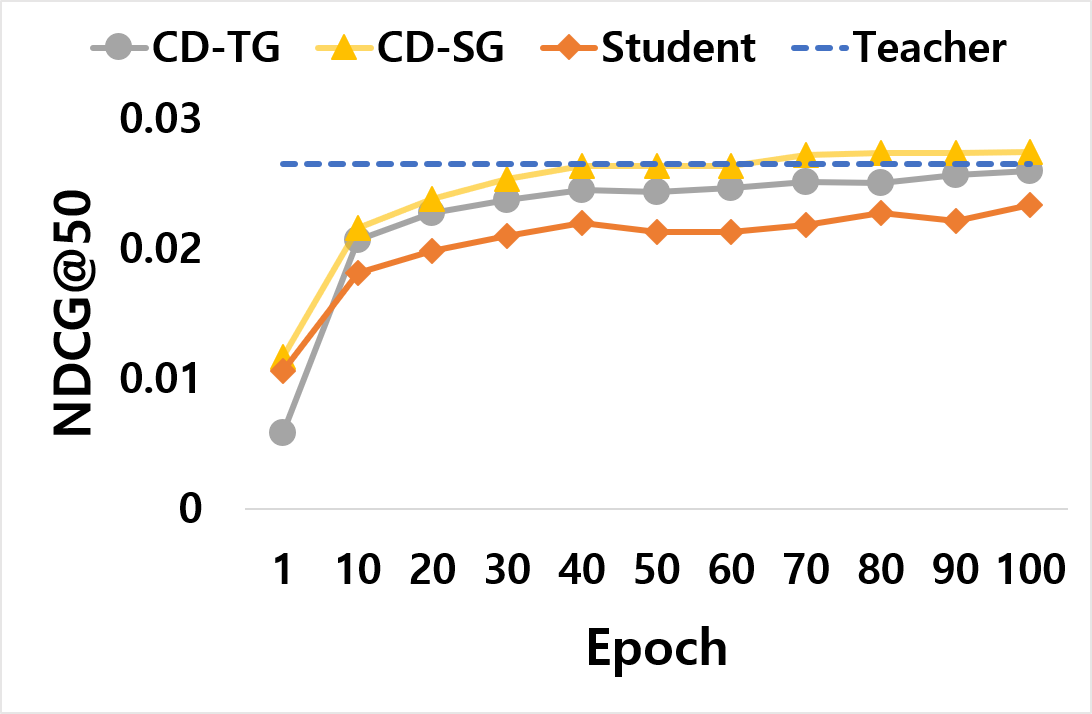} \\
(a) AMusic (CDAE) & (b) AMusic (Caser) & (c) Yelp (CDAE) & (d) Yelp (Caser) \\
\end{tabular}
\caption{NDCG@50 vs. the number of epochs. As the epoch increases, the performance gap between our model and teacher model is reduced.  }\label{fig:iterations}
\end{figure*}

For all KD models, the hyperparameters $\rho$ and $\lambda$ were controlled to properly reflect both CF and KD loss functions. For RD and RD-Rank, we used the public implementation of RD\footnote{https://github.com/graytowne/rank\_distill}. Also, the values of most hyperparameters were equal to their default values in public implementation. Note that there is a difference between $\lambda$ that appears in RD and CD. Specifically, we used the following parameter settings.

\begin{itemize}
    \item \textbf{RD}~\cite{TangW18b} and \textbf{RD-Rank}: We set $\rho$ to be 0.5. For CDAE, the number of items in the soft target was 15. For Caser, the number of items in the soft target was 10.

    \item \textbf{CD-Base}, \textbf{CD-TG} and \textbf{CD-SG}: We set $\lambda$ to be 0.5. For CDAE, we set $T_{1}$ to be 2 and $T_{2}$ to be 1. For ML100k, we set sampling size $K$ to be 0.8*$\mathcal|{I}_{u}|$. For other datasets, we set K to be 0.5*$\mathcal|{I}_{u}|$. For Caser, we set $T_{1}$ to be 1, $T_{2}$ to be 0, and sampling size K to be 50.
\end{itemize}

\vspace{2mm}
\noindent
\textbf{Environments}. We implemented our model and baseline models using TensorFlow 1.9.0 (CDAE) and PyTorch 1.0.0 (Caser). For Caser, we used the public PyTorch implementation~\footnote{https://github.com/graytowne/caser\_pytorch} provided in \cite{TangW18b}. All experiments were conducted on a desktop with 128 GB memory and 2 Intel Xeon Processor E5-2630 v4 (2.20 GHz, 25M cache), and all models were trained using 4 Nvidia GeForce GTX 1080Ti.

\subsection{Experimental Results}

\vspace{2mm}
\noindent
\textbf{Overall results}. Table~\ref{tab:com1} reports the performance of several variants of our model (\ie, CD-Base, CD-TG and CD-SG) and baseline models (\ie, RD, and RD-Rank). Among the four benchmark datasets, we compare KD models with the two baseline CF models. Teacher and student models indicate the baseline CF model with different parameters without KD. Also, the gain indicates how additional accuracy achieved by the proposed model over that of RD~\cite{TangW18b}.

Based on this evaluation, we found several interesting observations. Firstly, both CD-TG and CD-SG significantly outperform RD over all datasets. Note that the improvement gap for RD is somewhat different from that in ~\cite{TangW18b}. It is because we used leave-one-out evaluation while~\cite{TangW18b} used cross-validation evaluation. Our models are consistently better than RD by 2.7--33.2\% and 2.7--29.1\% in HR and NDCG, respectively. Also, CD-Base achieves better accuracy than RD. In this sense, our solution improves the CF loss function and helps boost the performance of \topN\ recommendation.

Secondly, CD-TG and CD-SG mostly achieve better accuracy than CD-Base. This implies that the rank-aware sampling method is more appropriate for addressing the ranking problem. However, RD-Rank tends to be comparable or slightly worse than RD. We conjecture that this is because RD possesses a prediction bias toward negative feedback in the CF loss function. Since selecting top-$K$ items in RD inherently induces the bias toward positive feedback, the bias in the original CF loss function helps to mitigate the sampling bias in RD. For this reason, RD-Rank is not as effective as our models. As the CF loss function in CD is not influenced by missing feedback, unlike RD, our models do not compensate for the negative bias by introducing a positive bias. As a result, our sampling strategy in the KD loss function is useful for boosting the prediction accuracy.

Lastly, our models consistently show improvements for two CF models with different architectures; while CDAE is based on the autoencoder for the offline recommendation, Caser is based on convolutional neural networks for a session-based recommendation. In particular, our models are most effective
in AMusic. This dataset is relatively more sparse than the
other datasets, implying that our models effectively overcome the data sparsity problem. Based on these results, we prove that our models can be extended to various CF models as model-agnostic solutions for CF.

\vspace{2mm}
\noindent
\textbf{Effects of model size}. We evaluate the effect of the size of the student model. Fig.~\ref{fig:modelvseffectiveness} depicts the relationship between model size and efficiency. The model size is proportional to the accuracy of our model, as observed in~\cite{TangW18b} as well. The same tendency consistently holds in different CF models. In both CF models, ones of the small size perform comparably to the teacher model, where the model size is about 20\% of the teacher model.

We further investigate the trade-off between model size and inference time. As depicted in Fig.~\ref{fig:modelvsinference}, as the model size increases, it requires greater amounts of inference time. This is reasonable because many model parameters require a higher computational cost. Compared to the teacher model, our models require less than about 50\% of the inference time, even though they achieve comparable performances to the teacher model. It can be concluded that our models are capable of transcending the trade-off between effectiveness and efficiency.

Lastly, Fig.~\ref{fig:iterations} depicts the effects of CD-TG and CD-SG on each training step. Our models consistently outperform the student model at each step. The number of training steps is inversely proportional to the performance gap between our models and the teacher model. As depicted in Fig.~\ref{fig:iterations}(d), CD-SG outperforms the teacher model even after 70 epochs.

\begin{figure}[t]
\centering
\begin{tabular}{cc}
\includegraphics[width=0.23\textwidth]{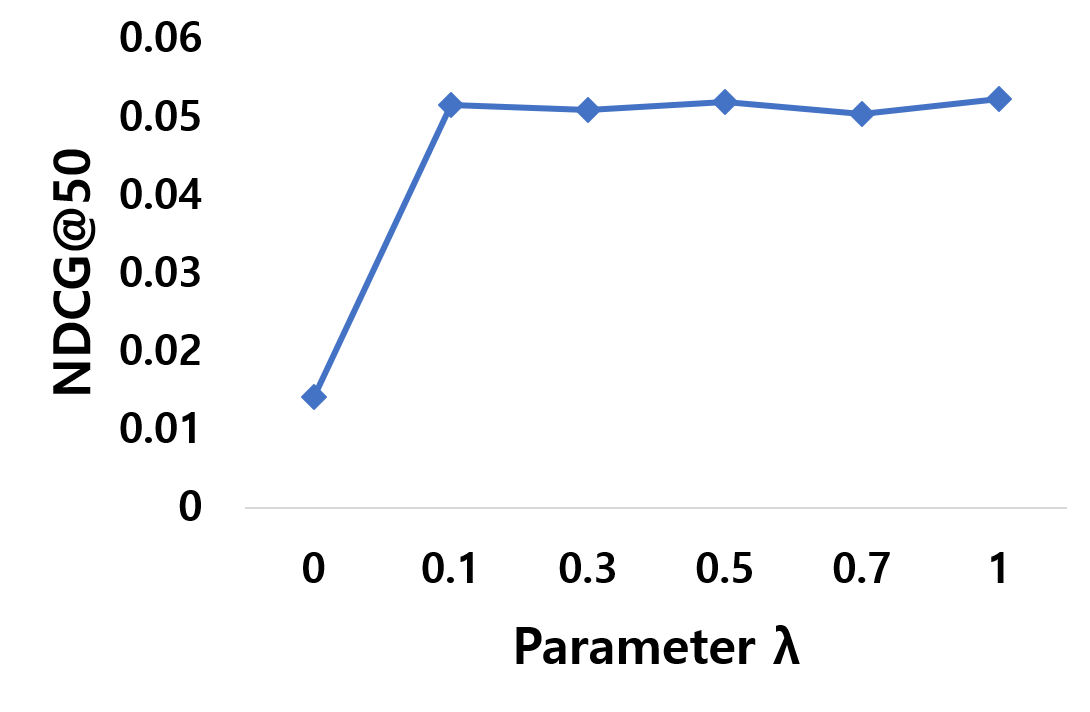} &
\includegraphics[width=0.23\textwidth]{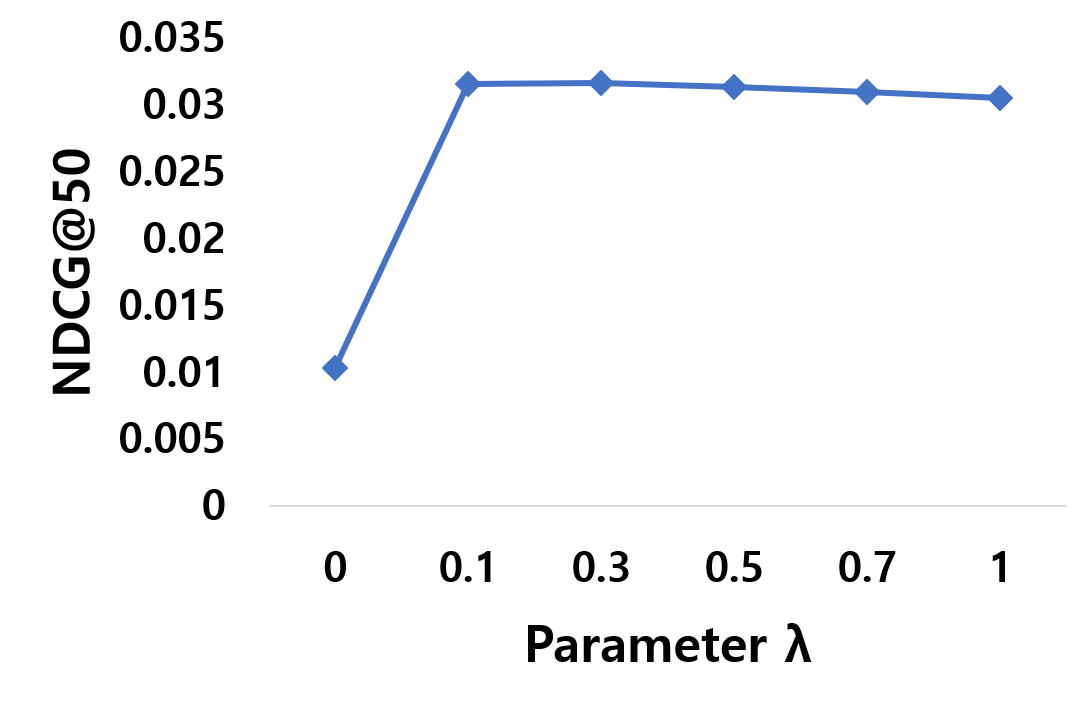} \\
(a) AMusic (CD-SG) & (b) Yelp (CD-SG) \\
\end{tabular}
\caption{NDCG@50 vs. KD parameter $\lambda$ (CDAE). In both datasets, our model is the best in $\lambda = 0.1$. }\label{fig:kdparameter}
\end{figure}

\begin{figure}[t]
\centering
\begin{tabular}{cc}
\includegraphics[width=0.23\textwidth]{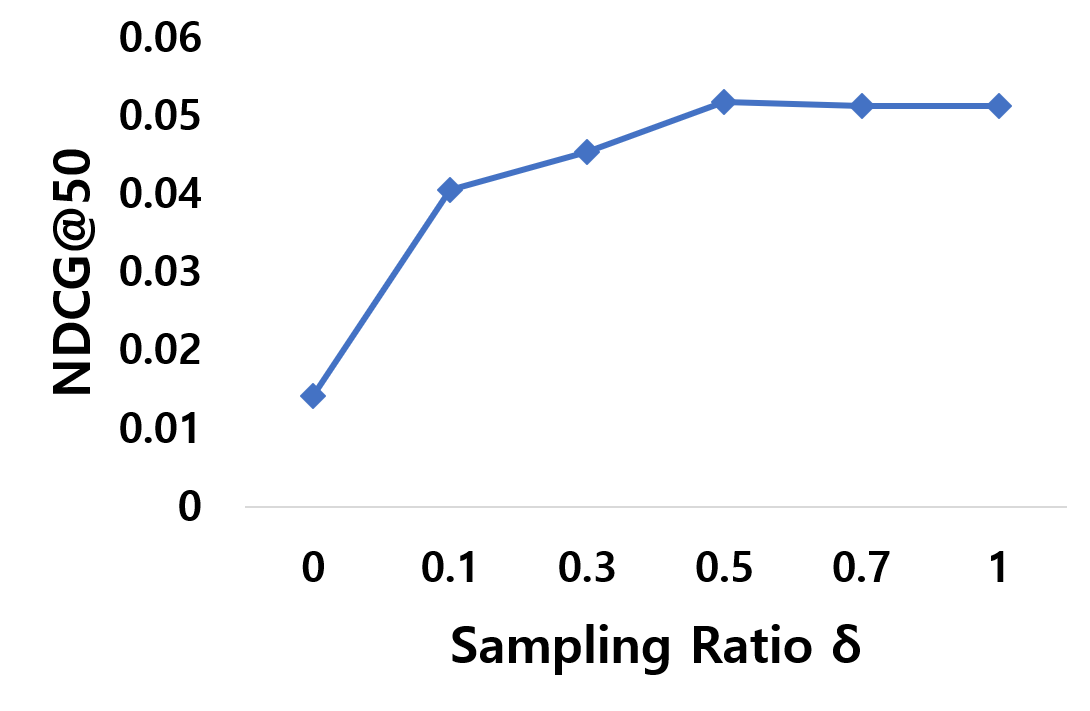} &
\includegraphics[width=0.23\textwidth]{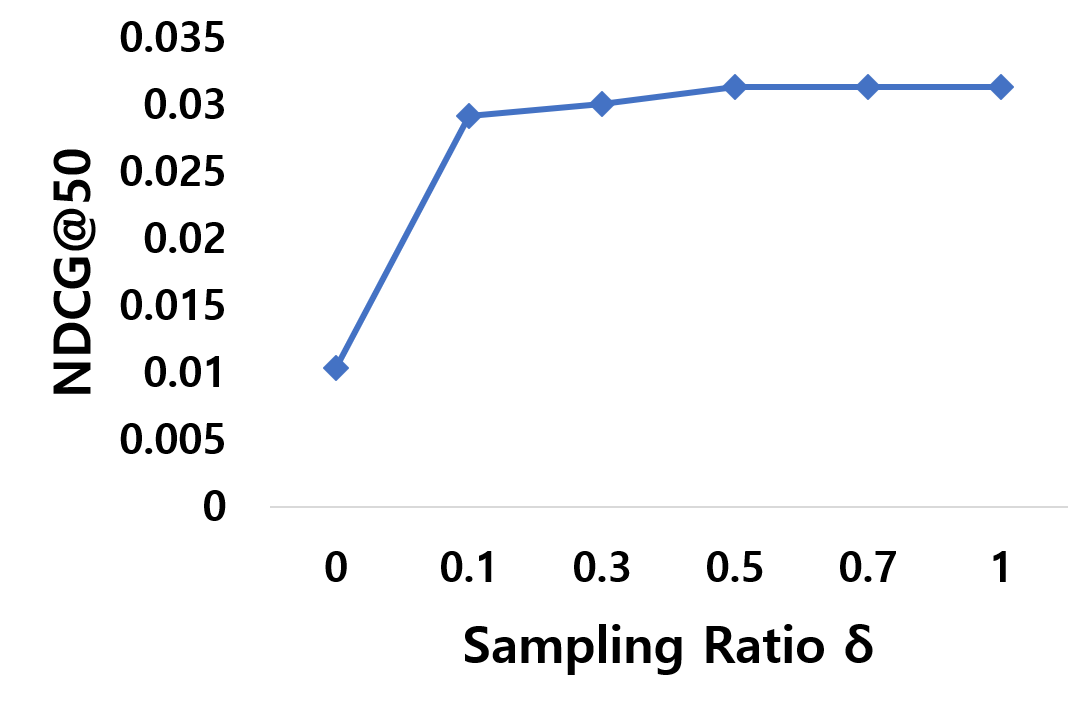} \\
(a) AMusic (CD-SG) & (b) Yelp (CD-SG) \\
\end{tabular}
\caption{NDCG@50 vs. sampling ratio $\delta$ (CDAE).  In both datasets, our model is the best in $\delta = 0.5$.}\label{fig:samplingratio}
\end{figure}

\vspace{2mm}
\noindent
\textbf{Effect of model hyperparameters}. Fig.~\ref{fig:kdparameter} depicts NDCG@50 over varying hyperparameter $\lambda$ for the KD loss function. In case of CD-SG, the best performance is achieved when $\lambda$ is around 0.1 in both datasets. Fig.~\ref{fig:samplingratio} depicts the NDCG for various sampling ratios. For CDAE, we sample $\delta \times |\mathcal{I}_{u}^{-}|$ items from those with missing feedback. In both datasets, the best performance is achieved when $\delta$ is around 0.5 for CD-SG.For CD-TG, we also observed a similar tendency for $\lambda$. 

For the sampling ratios $\delta$, CD-TG is worse than CD-SG when $\delta$ is approximately 0.1--0.5. This implies that CD-SG is more effective than CD-TG when the sampling ratio is low. In this sense, CD-SG is advantageous owing to its nature of leading a better update path for low sampling ratios.

\section{Related Work}\label{sec:relatedwork}

\vspace{2mm}
\noindent
\textbf{Model compression techniques}. Balancing the effectiveness and efficiency of computational models is an fundamental issue for real-world applications. To address this problem, various techniques~\cite{ChengWZZ17} have been widely developed to compress cumbersome models into smaller ones. In general, existing work falls into three categories: (1) \emph{binarization} and \emph{discretion}, (2) \emph{pruning and sharing} model parameters, and (3) \emph{knowledge distillation (KD)}.

First, \cite{CourbariauxBD15,HubaraCSEB16} proposed the binary encoding of model parameters. Under this method, real-valued model parameters are discretized via binary representation. Although the discretized model parameters incur the loss of accuracy, it can reduce the memory size and enhance efficiency. Second, the pruning and sharing method presented in~\cite{SrinivasB15,HanPTD15} removes or binds model parameters which are redundant or have minimal impacts in loss functions. In principle, these research directions focus on designing an efficient inference process using various computational acceleration techniques with low memory usage, thus mostly using model-dependent techniques.

Recently, KD is a model-independent learning framework that compresses a model by transferring the distilled knowledge of a large teacher model to a small student model. Various KD techniques have been proposed to improve the original KD toward two directions: (1) incorporating more information in addition to utilizing soft targets and (2) analyzing the loss function for KD.

The first trend is based on the intuition that the utilization of soft targets alone is not sufficient because meaningful intermediate information may be ignored during student training. FitNet~\cite{ref12} first pointed out such limitation and suggested using the output of intermediate layers as additional matching criteria. Similarly, \cite{ref13} utilized the gram matrix of the channel responses from the teacher model as additional information to educate the student model. Net2Net~\cite{ref24} employed model parameters of the teacher model directly to initialize those of the student model. Recently, \cite{ref03} used the attention map as an additional matching constraint. That is, in addition to the original loss term for matching the soft target, the attention map of the student model should match that of the teacher model. Most recently, \cite{ref11} further improved the attention-based method by matching the gradients (\ie, Jacobians) of output activations for the input.

Along an alternative direction, several algorithms focused on analyzing the choice of loss functions for KD. \cite{ref23} observed that the distance-based loss is inappropriate for transferring activation boundaries, and thus suggested a hinge loss. \cite{ref09} and \cite{ref10} employed adversarial learning into the KD framework. Recently, KDGAN \cite{ref04} bypassed the convergence step of adversarial learning by employing a triple-player game \cite{ref26}. In this study, we develop an improved loss function for KD. Unlike existing models, we mainly focus on modifying the loss function of KD for \topN\ recommendation.

\vspace{2mm}
\noindent
\textbf{One-class collaborative filtering (OCCF)}. For implicit datasets, handling missing feedback that intrinsically delineates a mixture of positive/negative feedback is a non-trivial issue. To address this challenge, existing studies can be categorized into \emph{weight-based}, \emph{sampling-based}, and \emph{imputation-based} methods. First, the weight-based method~\cite{HuKV08,PanS09} regards all missing feedback as negative ones. For instance, Hu et al.~\cite{HuKV08} and Pan et al.~\cite{PanS09} controlled weights as the confidence of negative values with various schemes, such as uniform, user-oriented, and item-oriented methods. Second, Paquet and Koenigstein~\cite{PaquetK13} proposed a sampling-based method by considering the degree distributions of users/items in the graph. Lastly, Sindhwani et al.~\cite{SindhwaniBHM10} regarded unobserved feedback as optimization variables and imputed missing feedback via optimization. Besides, Li et al.~\cite{LiHZC10} leveraged side information to construct user-item similarity, and Zheng et al.~\cite{ZhengDMZ13} employed multiple similarity matrices between users and items to predict drug-target interaction. Moreover, Yao et al.~\cite{YaoTYXZSL14} proposed dual regularization by combining the weighted- and imputation-based methods.

The proposed model is similar to the imputation method because the student model utilizes some inferred values for missing feedback. While the imputation-based method mainly focused on substituting missing feedback to improve the accuracy of \topN\ recommendation using auxiliary information, our model aims to balance the effectiveness and efficiency of \topN\ recommendation in the paradigm of KD.
\section{Conclusion}\label{sec:conclusion}

In this study, we propose a new knowledge distillation model, namely collaborative distillation (CD), with implicit user feedback for \topN\ recommendation. Specifically, we address several challenges raised in \topN\ recommendation: data sparsity, the ambiguity of missing feedback, and the ranking problem. To overcome these challenges, we first introduce a new loss function for CF. Then, we deal with the ranking problem using the rank-aware sampling method. In this process, our model utilizes the soft target without manipulation to manage data sparsity. Furthermore, we devise teacher- and student-guided training strategies to validate how the active/passive interactions between teacher and student models affect KD performance. Through extensive experiments, we demonstrate that the proposed model significantly outperforms the state-of-the-art model and several baseline models.

\section*{Acknowledgment}

This work was supported by the National Research Foundation of Korea (NRF) grant (No. NRF-2018R1A2B6009135 and NRF-2019R1A2C2006123) and the Institute of Information \& communications Technology Planning \& Evaluation(IITP) grant funded by the Korea government (MSIT) (No.2019-0-00421, AI Graduate School Support Program).

\bibliographystyle{IEEEtran}
\bibliography{references}


\end{document}